\documentclass{article}
\usepackage{ijcai25}

\usepackage{times}
\usepackage{soul}
\usepackage{url}
\usepackage[hidelinks]{hyperref}
\usepackage[utf8]{inputenc}
\usepackage[small]{caption}
\usepackage{graphicx}
\usepackage{amsmath,amsfonts}
\usepackage{amsthm}
\usepackage{booktabs}
\usepackage{algorithm}
\usepackage{algorithmic}
\usepackage[switch]{lineno}

\usepackage{array}
\usepackage[caption=false,font=normalsize,labelfont=sf,textfont=sf]{subfig}
\usepackage{textcomp}
\usepackage{verbatim}
\usepackage{booktabs}
\usepackage{multirow}
\usepackage{enumitem}
\usepackage{threeparttable}
\usepackage{colortbl}
\usepackage[capitalize]{cleveref}
\usepackage{subcaption}
\def\eg{\emph{e.g}\@addpunct{.}}
\def\eg{\emph{e.g}\@addpunct{.}}
\def\Eg{\emph{E.g}\@addpunct{.}}
\def\ie{\emph{i.e}\@addpunct{.}} 
\def\Ie{\emph{I.e}\@addpunct{.}}
\def\cf{\emph{cf}\@addpunct{.}} 
\def\Cf{\emph{Cf}\@addpunct{.}}
\def\etc{\emph{etc}\@addpunct{.}} 
\def\vs{\emph{vs}\@addpunct{.}}
\def\wrt{w.r.t\@addpunct{.}} 
\def\dof{d.o.f\@addpunct{.}}
\def\iid{i.i.d\@addpunct{.}} 
\def\wolog{w.l.o.g\@addpunct{.}}
\def\etal{\emph{et al}\@addpunct{.}}

\definecolor{mygray2}{gray}{.6}
\definecolor{mygray3}{gray}{.3}

\title{Denoise-then-Retrieve: Text-Conditioned Video Denoising for \\ Video Moment Retrieval}
\author{
Weijia Liu$^1$\and
Jiuxin Cao$^1$\and
Bo Miao$^2$\and
Zhiheng Fu$^3$\and
Xuelin Zhu$^3$\and
Jiawei Ge$^1$\and
Bo Liu$^1$\and
Mehwish Nasim$^4$\And
Ajmal Mian$^4$\\
\affiliations
$^1$Southeast University\\
$^2$The University of Adelaide\\
$^3$The Hong Kong Polytechnic University\\
$^4$The University of Western Australia\\
\emails
\{weijia-liu, jx.cao\}@seu.edu.cn
}
\begin{document}

% \crefname{section}{Sec.}{Secs.}
% \Crefname{section}{Section}{Sections}
% \crefname{figure}{Figure}{Figures}
% \crefname{table}{Table}{Tables}
% \Crefname{table}{Table}{Tables}

\maketitle

\begin{abstract}
% Current text-driven Video Moment Retrieval (VMR) methods use all video clips for multimodal encoding, even though many are clearly irrelevant to the text. These irrelevant noisy clips disrupt multimodal alignment and hamper model optimization.
Current text-driven Video Moment Retrieval (VMR) methods encode all video clips, including irrelevant ones, disrupting multimodal alignment and hindering optimization.
To this end, we propose a denoise-then-retrieve paradigm that explicitly filters text-irrelevant clips from videos and then retrieves the target moment using purified multimodal representations.
Following this paradigm, we introduce the Denoise-then-Retrieve Network (DRNet), comprising Text-Conditioned Denoising (TCD) and Text-Reconstruction Feedback (TRF) modules.
TCD integrates cross-attention and structured state space blocks to dynamically identify noisy clips and produce a noise mask to purify multimodal video representations.
TRF further distills a single query embedding from purified video representations and aligns it with the text embedding, serving as auxiliary supervision for denoising during training.
% TRF reconstructs the text embedding by distilling a query embedding from purified video representations and aligns it with the original text embedding, serving as auxiliary supervision for denoising during training.
Finally, we perform conditional retrieval using text embeddings on purified video representations for accurate VMR.
% Experiments on the Charades-STA and QVHighlights benchmarks show that our approach significantly outperforms existing state-of-the-art methods across all metrics. 
% Moreover, our denoise-then-retrieve paradigm is adaptable and can be seamlessly integrated into existing advanced VMR models to significantly enhance their performance.
Experiments on Charades-STA and QVHighlights demonstrate that our approach surpasses state-of-the-art methods on all metrics. Furthermore, our denoise-then-retrieve paradigm is adaptable and can be seamlessly integrated into advanced VMR models to boost performance.
% The Text-Reconstruction Feedback (TRF) then refines the denoising process by aligning the re-generated query embedding distilled from purified video representations with the language embedding. This provides feedback on the denoising quality. Moreover, our TRF improves moment retrieval by using enhanced query embeddings to guide the identification of target moments from masked video representations.
\end{abstract}

\section{Introduction}
Text-driven Video Moment Retrieval (VMR) aims to localize moments in untrimmed videos that semantically match the given query text. 
Unlike conventional temporal action localization constrained by predefined action categories, VMR flexibly localizes the moment through free-form linguistic expressions. 
VMR streamlines video analysis and benefits downstream applications~\cite{10572009,10486830}, including video semantic segmentation, user-friendly video editing, and mass surveillance. 

\begin{figure}[th]
    \centering
    \includegraphics[width=\linewidth]{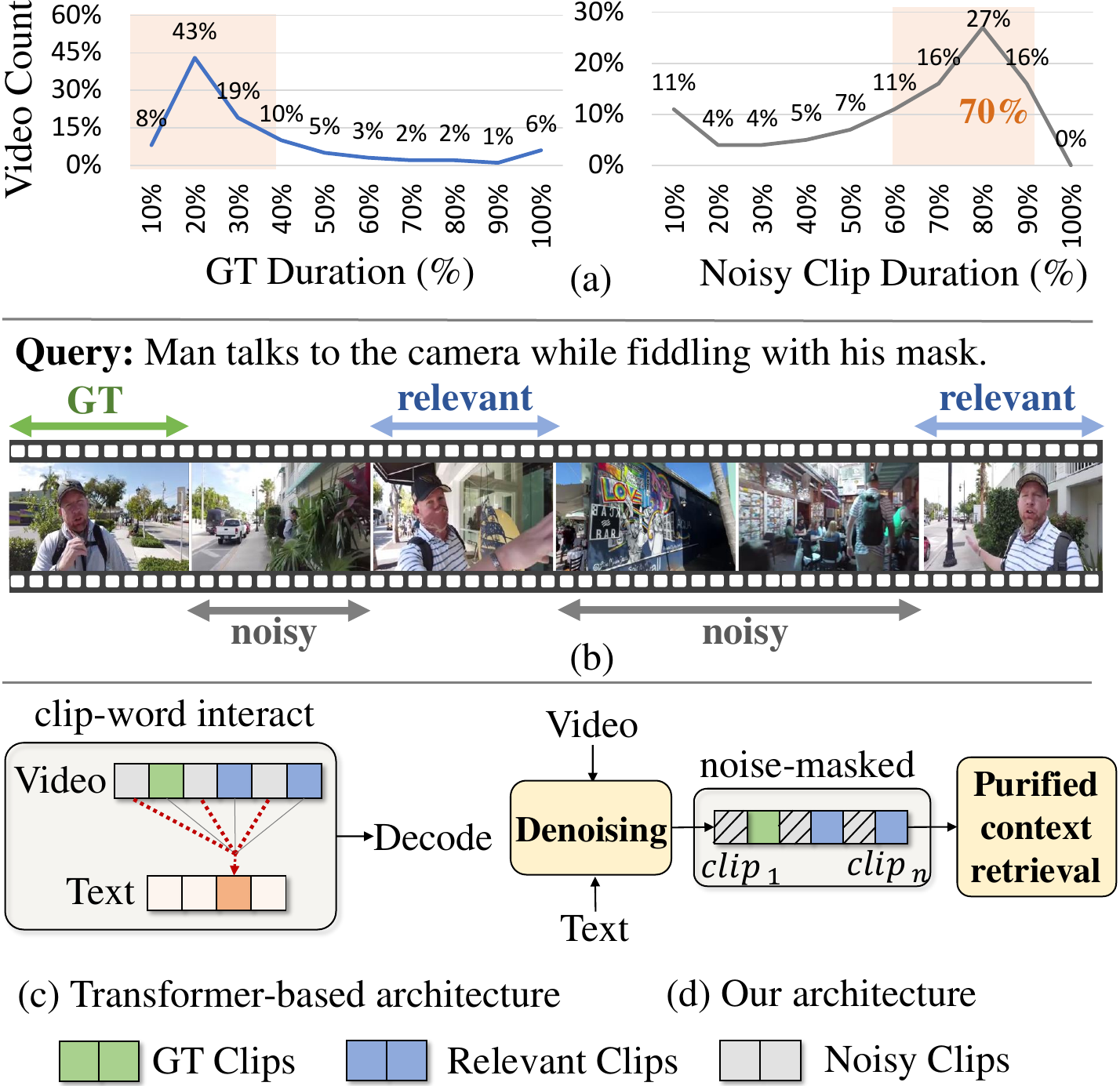}
    \caption{
    (a) Distribution of Ground Truth Moments (left) and noisy clips (right) relative to video duration in QVHighlights.
    (b) Our Text-Conditioned Denoising effectively identifies noisy and relevant video clips.
    % Noisy clips and relevant clips identified by our Text-Conditioned Denoising (TCD) module.
    (c) Previous transformer-based VMR methods use all video clips for multimodal encoding, including irrelevant ones.
    (d) Our DRNet approach explicitly excludes noisy video clips, enhancing purified multimodal modeling.
    }
    \label{intro}
    % \vspace{-3mm}
\end{figure}

Traditional two-stage VMR methods~\cite{xiao2021boundary,zhang2020span,xu2019multilevel} aim to extract a set of moment proposals using proposal generation networks, treating the task as a matching and ranking problem between the proposals and the query text. 
However, to achieve high recall, they generate numerous proposals of varying durations and locations, reducing efficiency and complicating matching

In recent years, Transformer architectures have become prevalent in multimodal understanding tasks~\cite{Miao_2023_ICCV,Miao_2024_NeurIPS}, including VMR~\cite{lin2023univtg,liu2022umt,moon2023query,xu2023mh,liu2024context}, due to their strong feature interaction and representation capabilities. These methods utilize transformer encoders to perform fine-grained word-clip interaction and integrate multimodal representations for moment retrieval.
However, they treat all video clips equally, inevitably introducing semantically irrelevant noisy clips into the multimodal representations (see \cref{intro} (c)) which eventually leads to suboptimal performance.

In VMR, input videos include both text-relevant and text-irrelevant clips.
Relevant clips are frames semantically aligned with the text query, including ground-truth frames (perfect alignment) and challenging non-target frames (partial alignment). Irrelevant (noise) clips are frames with little or no alignment. 
For example, in \cref{intro} (b), the clips where the “man talks to the camera" are text-relevant,
which can provide useful context for prediction.
In contrast, the scenery and walking clips are noise. 
We argue that relevant clips typically occupy only a small portion of a video, while noisy clips dominate. 
To validate this, we analyze all videos in the QVHighlights dataset.
%   by measuring the semantic similarities between the query text and each clip caption generated by ChatGPT
As shown in \cref{intro} (a), ground truth (GT) clips occupy less than 30\% of the duration in most videos, while noisy clips account for over 60\% clips in most videos. 
With excessive noisy clips throughout the video, generating abundant proposals with noisy clips or using all clips for transformer-based multimodal interaction can hinder the VMR model from focusing on the text-relevant clips that are more likely to be the target.

To address this issue, we propose a denoise-then-retrieve paradigm that explicitly removes noisy clips to narrow the retrieval range and strengthens purified multimodal representations for moment retrieval, as shown in~\cref{intro} (d).
% As shown in \cref{intro} (d) we propose explicit removal of the noisy clips to significantly narrow down the retrieval range, followed by extraction of the referred moments from the remaining smaller pool of relevant clips. 
Specifically, we design a Text-Conditioned Denoising (TCD) module to filter out noisy clips by dynamically generating noise masks. It integrates cross-attention and structured state space models for text-video interaction, and generates dynamic kernels to produce noise masks for purified multimodal representations.
To provide direct feedback on denoising quality, we introduce a Text-Reconstruction Feedback (TRF) module, which aligns the generated query from purified video features with the input text, serving as auxiliary supervision for TCD during training.
Finally, the decoder performs purified multimodal interaction between noise-masked video features and text embeddings, enabling accurate retrieval from text-relevant clips.
Additionally, when applied to other methods, our denoise-then-retrieve paradigm leads to notable performance improvements, showcasing the generalization capability. For example, UniVTG \cite{lin2023univtg} achieves an increase of 2.75\% points on the mAP@Avg metric. 
% As the re-generated query embeddings incorporate core visual semantics of relevant clips, analyzing the semantic differences between them and the input text can guide the MMR module to effectively distinguish the referred moment from the remaining relevant clips.
Our contributions are summarized as follows:
\begin{itemize}
    \item We propose the Denoise-then-Retrieve Network (DRNet) with Text-conditioned Denoising and Text-reconstruction Feedback. Our DRNet effectively extracts purified visual representations to enhance text-clip alignment, achieving top-tier performance on popular benchmarks.
    
    \item We propose a text-conditioned denoising approach that integrates cross-attention and structured state space blocks for effective multi-level multimodal fusion, generating dynamic kernels for accurate noise identification. 
    
    \item We introduce a text-reconstruction feedback mechanism that aligns the generated query from purified video features with the input text, providing auxiliary supervision for denoising during training.
    
    % To enhance noise filtering, We further devise a text-reconstruction feedback that enhances noisy clip filtering and moment retrieval using regenerated query embeddings from purified video representations.
    % It using the re-generated query embeddings from purified video representations to provide feedback on the denoising quality and guide the model to discriminate the target moments from the remaining text-relevant clips, respectively.

    \item  We demonstrate that the denoise-then-retrieve paradigm integrates seamlessly into current VMR models, yielding significant improvements across all metrics.
    % Integrating our paradigm seamlessly with current VMR models significantly enhances their performance on all metrics.
    % paradigm can adapt to other models.
\end{itemize}

Experiments on the Charades-STA and QVHighlights benchmarks show that our approach significantly outperforms existing state-of-the-art methods on all metrics. On Charades-STA, we surpass the nearest competitor MESM~\cite{liu2024towards} by 4.36\% points on the mAP@0.7 metric.

%\item We propose text-conditioned video denoising that integrates cross-attention into Mamba \cite{gu2023mamba} for multi-level multimodal fusion and creates dynamic kernels for noise identification, resulting in noisy clips filter.
%  also serves as a semantic supplement to the original text, aiding in the discrimination of target moments during localization.
% Our TCD performs multi-level text and video interaction by integrating cross-attention into Mamba\cite{gu2023mamba}, enhancing the model's multimodal comprehension and efficiency and creates dynamic kernels to discriminate noisy clips from the videos, generating noise masks.
% The resulting multimodal representation is used to adaptively generate a noise mask that excludes noisy clips from the video. 
% Meanwhile, Mamba has demonstrated excellent performance in the visual understanding domain \cite{gu2023mamba, zhu2024vision, chen2024video}. Its gating mechanism and linear computational complexity allow it to model long sequences effectively while maintaining good computational efficiency.
% Therefore, we propose a Text-Quided Video Denoising (QGVD) module that integrates cross-attention into Mamba to perform multi-level visual and textual fusion, enhancing the model's multimodal comprehension and efficiency. The resulting multimodal representation is used to adaptively generate a noise mask that excludes noisy clips from the video. 

\section{Related Work}

\noindent \textbf{Two-stage VMR Methods}~\cite{zhang2021multi,wang2021structured,chen2020learning,qu2020fine,yuan2019semantic} extract a set of moment proposals through multi-scale sliding windows or proposal-generating networks and treat the task as a matching and ranking problem between proposal candidates and the text query. However, sliding window-based methods~\cite{liu2018cross,jiang2019cross,ge2019mac} suffer from inefficient computation due to the re-computation of many overlapping areas in the densely sampled process with predefined multi-scale sliding windows. To reduce the number of candidates, proposal-based methods~\cite{xu2019multilevel,chen2019semantic,xiao2021boundary} devise various proposal-generating networks. For instance, QSPN~\cite{xu2019multilevel} generates proposals by introducing query representations as guidance for video encoding, while SAP~\cite{chen2019semantic} pre-trains a visual concept detection CNN with paired query-clip training data to calculate the visual-semantic correlation score for clips, grouping high-scoring clips to form proposals. Differently, BPNet~\cite{xiao2021boundary} directly utilizes VSLNet~\cite{zhang2020span} to generate moment proposals. To ensure high recall, these methods generate numerous proposals of varying durations and locations, which reduces efficiency and complicates matching.

% \noindent \textbf{Transformer-based VMR Approaches}~\cite{lin2023univtg, liu2022umt, moon2023query, lei2021detecting, xu2023mh, moon2023correlation, yang2024task} use Transformer encoders to perform word-clip level interactions between text and videos to establish a shared embedding space and regress the temporal span based on the aligned visual-text features. \cite{lei2021detecting} introduces detection Transformer (DETR) into the VMR task and models the task as a temporal moment detection problem. To fully exploit the information of a given query, \cite{moon2023query} uses cross-attention layers in the encoding stage to explicitly inject the context of text into video representation. 
% Unlike works that enforce text engagement in each clip, CG-DETR~\cite{moon2023correlation} carefully controls the degree of query text engagement in cross-modal interaction to enhance multimodal representations.
% However, text-referred moments occupy only a small portion of videos while noisy clips occupy a significant part and are spread throughout the video. As a result, treating all video clips equally in multimodal modeling can lead to suboptimal multimodal representations. In this work, we explicitly remove noisy clips to narrow the localization range and perform masked context aggregation and decoding to enhance moment retrieval.
\noindent \textbf{Transformer-based VMR Approaches}~\cite{lin2023univtg,liu2022umt,moon2023query,lei2021detecting,xu2023mh,moon2023correlation,yang2024task} use Transformer encoders to perform word-clip level interactions between text and videos to establish a shared embedding space and regress the temporal span based on the aligned visual-text features. \cite{lei2021detecting} introduces detection Transformer (DETR) into the VMR task and models the task as a temporal moment detection problem. To fully exploit the information of a given query, \cite{moon2023query} uses cross-attention layers in the encoding stage to explicitly inject the context of text into video representation. Unlike works that enforce text engagement in each clip, CG-DETR~\cite{moon2023correlation} carefully controls the degree of query text engagement in cross-modal interaction to enhance multimodal representations. However, text-referred moments occupy only a small portion of videos while noisy clips occupy a significant part and are spread throughout the video. As a result, treating all video clips equally in multimodal modeling can lead to suboptimal multimodal representations. In this work, we explicitly remove noisy clips to narrow localization range and perform masked context aggregation and decoding to enhance retrieval.
 
\section{Method}
\begin{figure*}[t]
    \centering
    \includegraphics[width=\linewidth]{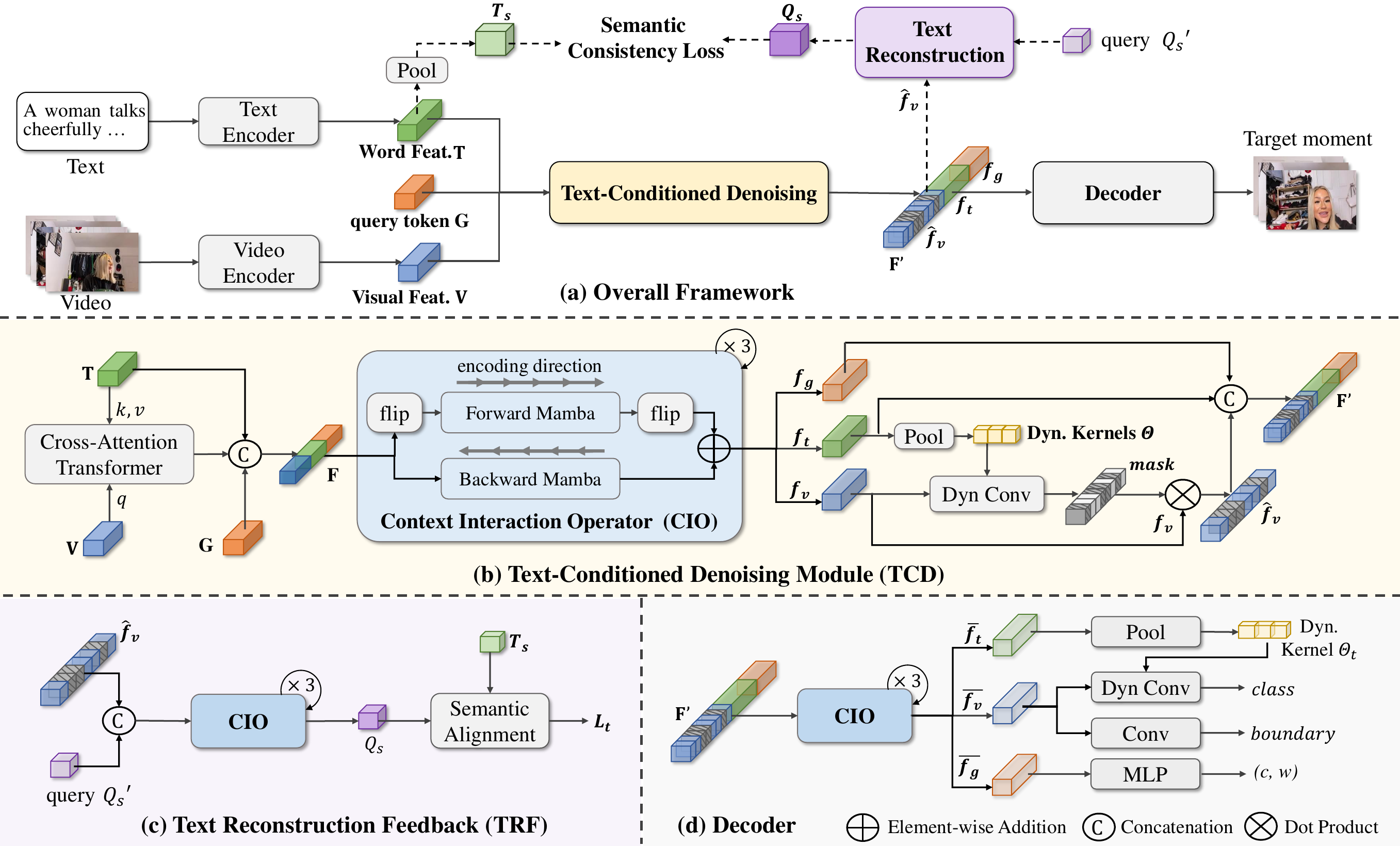}
    \caption{
    Overview of our DRNet. Dashed lines represent components used only during training. TCD identifies noisy clips and generates purified video representations by masking them. TRF provides feedback on TCD’s denoising quality using regenerated query embeddings from purified representations. The decoder performs multimodal interaction on the purified representations for accurate retrieval.
    }
    \label{framework}
    % \vspace{-4mm}
\end{figure*}

Given an untrimmed video $V$ containing ${L_v}$ clips and a text query $T$ with ${L_t}$ words, our objective is to localize the target moment as $(m_c, m_\sigma)$, where $m_c$ and $m_\sigma$ denote the central temporal coordinate and the span of the moment, respectively. \cref{framework} illustrates the overview of our DRNet and its modules.

Our overall architecture is described in \cref{framework}. Given a video and text representation extracted from fixed back-bones, DRNet first identifies noisy clips and generates purified video representations by masking them.
To further enhance the denoising, we regenerate query embeddings from purified video representations and aligns it with the input text, providing auxiliary supervision for denoising process during training. Finally, the decoder performs multimodal interaction on the purified representations for accurate retrieval.

\noindent \textbf{Video and Text Encoders.} 
We represent video features using the concatenation of CLIP~\cite{radford2021learning} and SlowFast~\cite{feichtenhofer2019slowfast}, and extract text features using the CLIP text encoder, consistent with previous works~\cite{lei2021detecting,liu2022umt,moon2023query}.
The input video $V$ and text $T$ are encoded with frozen encoders and projected to the same dimension $D$ via two Feed-Forward Networks (FFN).
The resulting video embeddings $\bold V = [v_1, v_2, ..., v_{L_v}]$ and text embeddings $\bold T = [t_1, t_2, ..., t_{L_t}]$ are then fed into the TCD module.

\subsection{Text-Conditioned Denoising (TCD)}\label{TCD}
As shown in \cref{framework} (b), to dynamically identify noisy clips, we leverage the query text to guide video representation learning, multimodal interaction, and noise mask generation. Specifically, we first inject textual context into video clips via cross-attention, obtaining text-aware video representations. These representations, along with text features, are then processed by state space models to propagate intra- and inter-modal context, generating refined multimodal representations. Finally, text-driven dynamic convolution kernels are constructed to identify and filter out noisy clips.
% o produce clip-wise representations equipped with information regarding the degree of query-relevance sinc

% Recent VMR methods use transformer-based architectures, however, these methods are susceptible to noisy clips due to their indiscriminate use of all tokens for multimodal modeling, and suffer from quadratic complexity.
% Recently, state space models such as Mamba~\cite{gu2023mamba} have excelled in visual understanding \cite{zhu2024vision, chen2024video}. Their gating mechanisms and linear complexity enable effective modeling of long sequences with high efficiency. 
% To leverage their strengths, we propose integrating cross-attention and Mamba to integrate text semantics into video clips and selectively propagate intra- and inter-modal context, generating refined multimodal representation for noise filtering.

We first apply cross-attention between video clips and text embeddings to enhance the target awareness. Here, we use italicized letters to denote the \textit{query}, \textit{key}, and \textit{value} in the cross-attention layer. Specifically, the video representations $\bold V$ serve as the \textit{query}, while the textual features $\bold T$ are used as the \textit{key} and \textit{value}:
\begin{equation}
    {\rm Attn}(Q_{\bold V}, K_{\bold T}, V_{\bold T}) = {\rm softmax}(\frac{Q_{\bold V} (K_{\bold T})^T}{\sqrt{D}})V_{\bold T}
\end{equation}
The updated video representations are obtained by computing a weighted sum of the text features $\bold T$, where the attention scores are projected through a Multi-layer Perceptron (MLP) and integrated into the original video representations, resulting in text-aware video representations $\hat{\bold V} = [\hat{v}_1, \hat{v}_2, ..., \hat{v}_{L_v}]$.

% We concatenate $\hat{\bold V}$, $\tilde{\bold T}$, and $\tilde{\bold G}$ to produce the multimodal sequence $\bold S$ for context integration. 

For text-clip interaction and context integration, we design the Context Interaction Operator (CIO) based on state space models, which have demonstrated significant success in visual understanding tasks\cite{zhu2024vision,chen2024video}, including Mamba~\cite{gu2023mamba}. The gating mechanisms and linear complexity of these models facilitate efficient modeling of long sequences. Specifically, each CIO consists of two separate Mamba blocks~\cite{gu2023mamba} that propagate the context of feature sequences in both forward and backward temporal directions.

We concatenate the text-aware video representations $\hat{\bold V}$ with the text features to form a multimodal sequence, and add learnable global tokens at the end of the sequence to aggregate global features. These global tokens are denoted as $\bold G = [g_1, g_2, ..., g_{L_q}]$, where $L_g$ denotes the number of global tokens.
To preserve positional and modality-specific information during cross-modal interaction, learnable position embeddings $\bold E^{p}$ and modality-type embeddings $\bold E^{m}$ are incorporated into each modality. The input multimodal sequence $\bold F$ to the CIOs is then represented as
\begin{gather}
    \tilde{\bold V} = \hat{\bold V} + \bold E^{p}_V + \bold E^{m}_V, \\
    \tilde{\bold T} = \bold T + \bold E^{p}_T + \bold E^{m}_T, \\
    \tilde{\bold G} = \bold G + \bold E^{p}_G + \bold E^{m}_G, \\
    \bold F = [\tilde{\bold T}, \tilde{\bold V}, \tilde{\bold G}]
\end{gather}
where $\bold F \in \mathbb{R}^{(L_g+L_t+L_v)\times D}$.
After encoding with the bi-directional Mambas in the CIO, the output of the backward Mamba block is flipped and added to the corresponding output of the forward Mamba block along the temporal channel, completing one round of context interaction. The multimodal sequence $\bold F$ undergoes intra- and inter-modal context integration by stacking three CIOs.

% The interactions occur at three levels:
% 1) Intra-modal contextual interaction: Tokens within each modality learn contextual semantics from surrounding tokens through forward and backward Mamba propagation.
% 2) Cross-modal contextual interaction: The forward Mamba block propagates textual semantics into the visual features, while the backward Mamba block integrates visual semantics into the textual features.
% 3) Global context integration: Through the information propagation of the bi-directional Mambas, the global tokens at the end of the sequence integrate semantics from both text and video features.
% By stacking multiple CIOs, we generate a context-aware multimodal sequence, where $\boldsymbol{f_t}$, $\boldsymbol{f_v}$, and $\boldsymbol{f_g}$ represent the text, video, and global features within the sequence, respectively.
The interactions occur at three levels:
1) Intra-modal contextual interaction: Tokens within each modality learn contextual semantics from surrounding tokens through forward and backward Mamba propagation.
2) Cross-modal contextual interaction: The forward Mamba block propagates textual semantics into visual features, while the backward Mamba block integrates visual semantics into textual features.
3) Global context integration: Through information propagation of bi-directional Mambas, the global tokens at the end of the sequence integrate semantics from both text and video features.
By stacking multiple CIOs, we generate a context-aware multimodal sequence, where $\boldsymbol{f_t}$, $\boldsymbol{f_v}$, and $\boldsymbol{f_g}$ represent text, video, and global features, respectively.

\textbf{Dynamic Denoising.}
To purify the visual context, we employ text features to create dynamic kernels~\cite{chen2020dynamic}, which performs point-wise convolutions to identify noisy clips. 
The word-level text feature $\boldsymbol{f_t}$ is first pooled to a fixed length of $L_k$  to handle varying text lengths, followed by a fully-connected layer to generate dynamic kernels $\Theta = \{ \theta_i \}_{i=1}^{N_k}$, where $N_k$ is the number of kernels, and $\theta_i \in \mathbb{R}^{D\times1}$. 
These dynamic kernels are then applied to $\boldsymbol{f_v}$ to update the visual features:
\begin{gather}
    \boldsymbol{f_v}' = (\varphi( \theta_1 \boldsymbol{f_v}  \oplus \dots \oplus \theta_{N_k} \boldsymbol{f_v}) + \boldsymbol{f_v}) 
\end{gather}
where $\oplus$ denotes concatenation along the channel dimension, and $\varphi(\cdot)$ represents a $1 \times 1$ convolution for dimensionality reduction.

After applying the Sigmoid function $\sigma$, we obtain the text-clip alignment scores $\bold{S} = [s_1, s_2, \ldots, s_n]$, which indicate the degree of semantic relevance to the text.
By applying a threshold $\mu$, we generate the noise mask vector $\bold{M} = [m_1, m_2, \ldots, m_n]$,
\begin{gather}
    \bold{S} = \sigma( \boldsymbol{f_v}')  \\ 
     m_i =  \begin{cases} 
                1 & \text{if } s_i > \mu \\
                0 & \text{otherwise}
            \end{cases}
            \quad \text{for } i = 1, 2, \ldots, n
\end{gather}
where $m_i=0$ means that the $i$-th clip is semantically irrelevant to the text (a noisy clip), and $m_i=1$ means relevant.
Finally, $\bold{M}$ is applied to mask out noisy clips and produce purified visual representations $\boldsymbol{\hat{f_v}}$.
% Finally, the generate noise mask $\bold{M}$ is used to mask out noisy clips to 
% exclude noisy clips by performing a dot product with the visual features $\boldsymbol{f_v}$, resulting in 
% generate purified visual representations $\boldsymbol{\hat {f_v}}$.
\cref{fig_score} visualizes the text-clip alignment scores $\bold{S}$ and the noise mask 
 $\bold{M}$ for a given case, demonstrating that our TCD module effectively identifies relevant clips and filters out noise. 
 The updated textual, visual, and global features are then concatenated as $\bold{F'}=[\boldsymbol{f_t}, \boldsymbol{\hat{f_v}}, \boldsymbol{f_g}]$ and fed into the decoder.

\subsection{Text-Reconstruction Feedback (TRF)} \label{TRF}
TRF further enhances denoising by regenerating query embeddings from the purified visual representations. It aligns the regenerated query with the input text in textual semantic space, providing auxiliary supervision for denoising process during training. This feedback not only improves denoising quality but also strengthens the purified video representations.

Specifically, as shown in \cref{framework} (c), we distill a query embedding from the purified video representations $\boldsymbol{\hat{f_v}}$ using context interaction operators (CIO) (see \cref{TCD}), projecting $\boldsymbol{\hat{f_v}}$ into the text semantic space. To achieve this, we introduce a learnable query embedding $\bold{Q_s'} \in \mathbb{R}^{D \times 1}$ and construct a mapping network $\mathcal{D}$ by stacking three CIOs.
The query embedding $\bold{Q_s'}$ interacts with the purified video representations $\boldsymbol{\hat{f_v}}$ to generate the reconstructed sentence-level embedding:
\begin{equation}
    \bold{\hat{Q_s}} = \mathcal{D}(\boldsymbol{\hat{f_v}}, \bold{Q_s'}) 
\end{equation}
where $\bold{\hat{Q_s}}\in \mathbb{R}^{D \times 1}$.
The sentence-level embedding of the input text, $\bold{T_s}$, is obtained by average pooling its word-level embeddings, $\bold T$. We then compute the semantic consistency loss $\mathcal{L}_t$ between $\bold{T_s}$ and $\bold{\hat{Q_s}}$ using cosine similarity:
\begin{equation}
    \mathcal{L}_{t} = \lambda_t(1 - \frac{\bold{T_s} \cdot \bold{\hat{Q_s}}}{||\bold{T_s}|| \ ||\bold{\hat{Q_s}}||}),
\end{equation}
where $\lambda_t (\lambda_t=2)$ is a hyperparameter balancing the loss terms. Since the regenerated query embeddings from text-relevant clips capture key visual semantics, minimizing $\mathcal{L}_{t}$ forces the purified visual representations to maximally reflect the input text semantics, enhancing the denoising process.

% \begin{figure}
%     \centering
%     \includegraphics[width=\linewidth]{figures/4.pdf}
%     \caption{Overview of the Text-Reconstruction Feedback (TRF) and Masked Moment Retrieval (MMR) modules.}
%     \label{generation}
% \end{figure}

\subsection{Decoder}  \label{retrieval}
After masking noisy clips, we perform multimodal interaction on the purified visual and textual features to capture fine-grained differences between text-relevant clips. The resulting purified multimodal representations are then decoded for accurate retrieval.

As shown in \cref{framework} (d), we use context interaction operators (CIO) (see \cref{TCD}) to build a multimodal encoder $\mathcal{E}$ with three layers. The encoder takes the noise-masked multimodal sequence $\bold{F'}$ from the TCD module as input.
Through cross-modal context encoding, the purified multimodal representation $\bold{F}=\mathcal{E}(\bold{F'})$ is generated and passed to different decoding heads for target moment retrieval. We denote the global, textual, and visual features within the $\bold{F}$ sequence as $\boldsymbol{\bar{f_g}}$, $\boldsymbol{\bar{f_t}}$, and $\boldsymbol{\bar{f_v}}$, respectively.

% \vspace{1mm}
\textbf{Moment retrieval.}
1) Global retrieval. We use the global features $\boldsymbol{\bar{f_g}}$ to directly regress the central temporal coordinate $m_c$ via an MLP, and regress the moment span $m_\sigma$ via a fully connected layer. The global localization loss combines an L1 loss and a generalized IoU loss $\mathcal{L}_{gIoU}(\cdot)$ following~\cite{moon2023query}.
\begin{equation}
    \mathcal{L}_g = \lambda_{L1}^g ||m - \hat{m}|| + \lambda_{iou}^g\mathcal{L}_{iou}(m, \hat{m}),
\end{equation}
where $m$ is the ground-truth moment and $\hat{m}$ is the corresponding prediction, each containing the center coordinate and span.
% where $m$ and $\hat{m}$ are ground-truth moment and its correspond prediction containing center coordinate and span.
 
\noindent2) Boundary prediction. Following \cite{lin2023univtg}, we apply three $1 \times 3$ Conv layers with $N_k$ filters and ReLU activation to the output $\boldsymbol{\bar{f_v}} \in \mathbb{R}^{L_v \times D}$ from the multimodal encoder $\mathcal{E}$. The final layer has two output channels, representing the left and right offsets $\hat{d_i}\in \mathbb{R}^{2 \times L_v}$ for each clip. 
The predicted boundaries $\hat{b}_i$ are then calculated, and the boundary loss $\mathcal{L}_b$, which includes smooth L1 and IoU losses, is used to supervise the predictions.
\begin{gather}
    \mathcal{L}_b = \lambda_{L1}^b \mathcal{L}_{SmoothL1}(\hat{d_i}, d_i) + \lambda_{iou}^b \mathcal{L}_{iou}(\hat{b_i}, b_i).
\end{gather}

\noindent 3) Text-conditioned Foreground Classification.
As illustrated in \cref{intro} (d), we pool the text features $\boldsymbol{\bar{f_t}}$ to a fixed length $N_k$ and apply a fully connected layer to generate convolution kernels $\Theta_t \in \mathbb{R}^{D \times N_k}$.
The decoder head for clip classification (foreground/background) is the same as the boundary prediction head. However, in this case, we use $\Theta_t$ as the convolution kernels in the first layer, and the final layer outputs a single channel to classify each clip $\hat{c_i}$. Binary cross-entropy loss is applied for classification.
\begin{equation}
    \mathcal{L}_c = -\lambda_c(c_i log\hat{c_i} + (1-c_i)log(1-\hat{c_i})).
\end{equation}

\noindent 4) Contrastive Learning.
Following prior works~\cite{lei2021detecting,lin2023univtg,moon2023query}, we incorporate intra-video and inter-video contrastive learning during training. Intra-video contrastive learning treats clips within the ground truth moment as positive pairs and those outside as negative pairs, while inter-video contrastive learning uses text from other samples in the batch as negative pairs. The relevance between a clip embedding and a sentence-level text embedding $\bold{Q_s}$ is quantified by cosine similarity $r_i$.
\begin{gather}
    \mathcal{L}_r^{intra} = -log\frac{\text{exp}(r_p/\tau)}{\text{exp}(r_p/\tau) + \sum_{j\in \Omega} \text{exp}(r_j/\tau)} \\
    \mathcal{L}_r^{inter} = -log\frac{\text{exp}(r_p/\tau)}{\sum_{k\in \Omega'} \text{exp}(r_k/\tau)} ,
\end{gather}
where $\Omega$ and $\Omega'$ are negative sets, $r_p$ is the relevance score of positive samples, and $\tau$ is a temperature parameter. The overall contrastive learning loss is $\mathcal{L}_r = \lambda_{intra}\mathcal{L}_r^{intra} + \lambda_{inter}\mathcal{L}_r^{inter}$. 
Finally, after combining the textual reconstruction loss $\mathcal{L}_{t}$, our total training objective becomes:
\begin{equation}
    \mathcal{L} = \frac{1}{L_v}\sum_{i=1}^{L_v}(\mathcal{L}_r + \mathcal{L}_b + \mathcal{L}_c) + \mathcal{L}_g + \mathcal{L}_{t}.
\end{equation}
% where $\lambda_r, \lambda_b, \lambda_c, \lambda_g$, and $\lambda_t$ are weights to balance losses. 

\section{Experiments}
% \vspace{-20pt}
\begin{table}[t]
\centering
\renewcommand{\arraystretch}{1.15}
    \addtolength{\tabcolsep}{0pt}
    \resizebox{0.48\textwidth}{!}{
    \begin{tabular}{l|cccccc}
    \hline
          \multirow{2}{*}{Model} &  \multicolumn{2}{c}{R1} & \multicolumn{3}{c}{mAP} \\
          ~ & @0.5 & @0.7 & @0.5 & @0.75 & @Avg. \\
    \hline
         % MCN\textsubscript{17eccv} &  11.41 & 2.72 & 24.94 & 8.22 & 10.67\\
         % CAL\textsubscript{17eccv} &  25.49 & 11.54 & 23.4 & 7.65 & 9.89\\
         % XML\textsubscript{20eccv} &  41.83 & 30.35 & 44.63 & 31.73 & 32.14\\
         % XML+\textsubscript{20eccv} &  46.69 & 33.46 & 47.89 & 34.67 & 34.9\\
         MDETR\textsubscript{21neurips} &  52.89 & 33.02 & 54.82 & 29.4 & 30.73\\
         UMT\textsubscript{22cvpr} &  56.23 & 41.18 & 53.38 & 37.01 & 36.12\\
         MomentDiff\textsubscript{24neurips} &  57.42 & 39.66 & 54.02 & 35.73 & 35.95\\
         UniVTG\textsubscript{23iccv} &  58.86 & 40.86 & 57.60 & 35.59 & 35.47\\ 
         QD-DETR\textsubscript{23cvpr} &  62.4 & 44.98 & 62.52 & 39.88 & 39.86\\
         % SFABD~\cite{huang2024semantic} &  - & - & 62.38 & 44.39 & 43.79\\
         MESM\textsubscript{24aaai} & 62.78 & 45.2 & 62.64 & 41.45 & 40.68\\
         UVCOM\textsubscript{24cvpr} &  63.55 & 47.47 & 63.37 & 42.67 & 43.18\\ 
         % LMR~\cite{liu2024context}  & 64.40 & 47.21 & \textbf{64.65} & 43.16 & 42.56\\
         TR-DETR\textsubscript{24aaai} &  64.66 & 48.96 & 63.98 & 43.73 & 42.62\\  
         % CG-DETR~\cite{moon2023correlation} &  65.4 & 48.4 & \textbf{64.5} & 42.8 & 42.9\\  
    \hline               
         \rowcolor[gray]{0.9} \textbf{Our Model}  & \textbf{66.73} & \textbf{50.52} & \textbf{64.17} & \textbf{45.79} & \textbf{43.73}\\
         % \hline
          % UnLoc-B~\cite{yan2023unloc}~\pub{ICCV23} & 64.50 & 48.80 & - & - & -\\
    \bottomrule
    \end{tabular}
    }
    \caption{Comparison on the QVHighlights \textit{test} split obtained from the official server. All methods use only video (no audio) data, with Slowfast and CLIP as the visual backbones for fair comparison.} 
    \label{QVHighlights}
\end{table}

% \vspace{-10pt}
\begin{table*}[t]
\centering
\renewcommand{\arraystretch}{1.15}
    \setlength{\tabcolsep}{5.5pt}
    \resizebox{\textwidth}{!}{%
     \begin{tabular}{l|cccccl|lccccc}
     \toprule
         \centering Method & feat. & R1@0.5 & R1@0.7 & mAP@0.5 & mAP@0.7 & mAP@Avg 
         & \centering Method & feat. & R1@0.5 & R1@0.7 & mIoU \\
    \hline
            % 2D-TAN\textsubscript{20aaai} & VGG & 41.34 & 23.91 & 54.68 & 24.15 & 29.26
            % & 2D-TAN\textsubscript{20aaai} & SF+C & 46.02 & 27.5 & 41.25 \\
             
            % MAN\textsubscript{19cvpr} & VGG & 41.24 & 20.54  & - & - & -
            % & VSLNet\textsubscript{20arxiv} & SF+C & 42.69 & 24.14 & 41.58\\
            
            RaNet\textsubscript{21arxiv}$\star$ &VGG & 42.91 &25.82 &53.28 &24.41 &28.55
            & MDETR\textsubscript{21neurips} & SF+C & 52.07 & 30.59 & 45.54 \\                
            
            MomentDETR\textsubscript{21neurips}$\star$ &VGG& 50.54 &28.01 &57.39& 25.62& 29.87
            & QD-DETR\textsubscript{23cvpr} & SF+C & 57.31 & 32.55 & - \\
                         
            UMT\textsubscript{22cvpr}$\ddagger$ & VGG & 48.44 & 29.76 & 58.03 & 27.46 & 30.37
            & VMS\textsubscript{24arxiv} & SF+C & 57.18 & 36.05 & -\\
             
            MMN\textsubscript{22aaai}$\star$ &VGG& 46.93 & 27.07& 58.85&  28.16&  31.58
            & UniVTG\textsubscript{23iccv} & SF+C & 58.01 & 35.65 & 50.1\\
                         
            QD-DETR\textsubscript{23cvpr}$\star$ & VGG & 51.51 & 32.69 & 62.88 & 32.6 & 34.46
            & TR-DETR\textsubscript{24aaai} & SF+C & 57.61 & 33.52 & -\\      
             
            MomentDiff\textsubscript{24neurips} & VGG & 51.94 & 28.25 & 59.86 & 29.11 & 31.66
            & LLMEPET\textsubscript{24arxiv}& SF+C & - & 36.49 & 50.25\\     
            
            MESM\textsubscript{24aaai} & VGG & 56.69 & 35.99  & 67.94 & 33.64 & 37.33
            & UVCOM\textsubscript{24cvpr} & SF+C & 59.25 & 36.63 & - \\
        \hline
            \rowcolor[gray]{0.9}  
            \textbf{DRNet} & VGG & \textbf{59.03} & \textbf{36.26} & \textbf{69.75} & \textbf{38} & \textbf{39.33}
            & \textbf{DRNet} & SF+C & \textbf{60.86} & \textbf{39.78} & \textbf{52.07}\\                         
    \bottomrule
    \end{tabular}
    }
    \caption{Comparison on Charades-STA \textit{test} split. $\ddagger$: methods that use additional audio data. $\star$: results re-implemented under the same training strategies as \protect\cite{li2024momentdiff,liu2024towards}. SF+C: SlowFast and CLIP features.}
    \label{Charades}
\end{table*}

\begin{table*}[t]
\centering
\renewcommand{\arraystretch}{1.15}
    \addtolength{\tabcolsep}{0pt}
    \resizebox{0.6\textwidth}{!}{%
    \begin{tabular}{c|ccc|c|c|ccccc}
    \hline
        \multirow{2}{*}{~} & \multicolumn{3}{c|}{\textbf{TCD}} & \multirow{2}{*}{\textbf{TRF}} &\multirow{2}{*}{\textbf{Decoder}} & \multicolumn{2}{c}{\textbf{R1}} & \multicolumn{3}{c}{\textbf{mAP}} \\ 
        % \cline{2-5} \cline{8-12}
         ~  & CA & DK & LGT      & ~ & ~       & @0.5 &  @0.7 & @0.5 & @0.75 & Avg. \\
    \hline
         A1 & ~  & ~ & ~      & $\checkmark$ & $\checkmark$     & 56.45 & 39.61 & 56.55 & 35.94 & 34.89\\      
         A2 & $\checkmark$ & $\checkmark$ & $\checkmark$  & ~ & $\checkmark$    & 66.84 & 51.87 & 65.23 & 46.79 & 45.18\\
         A3 & $\checkmark$ & $\checkmark$ & $\checkmark$  & $\checkmark$ & ~ &  67.21 & 52.63 & 64.87 & 46.52 & 44.82\\
         % & 65.94 & 51.42 & 64.06 & 45.7 & 44.18\\
         A4 & $\checkmark$ & $\checkmark$ & $\checkmark$  & ~ & ~      & 66.84 & 51.23 &65.1& 46.19 & 44.51\\
    \hline
         A5 &  \multicolumn{5}{c|}{Mamba $\rightarrow$ Transformer }   & 64.58 & 47.74 & 61.67 & 42.45 & 40.07\\
    \hline
         B1 & ~ & $\checkmark$ & $\checkmark$ & $\checkmark$ & $\checkmark$  & 63.29 & 49.68 &61.48& 43.76 & 42.62\\      
         B2 & $\checkmark$ & ~ & $\checkmark$ & $\checkmark$ & $\checkmark$  & 67.81 & 52.97 & 64.51 & 46.6 & 44.98\\
         B3 & $\checkmark$ & $\checkmark$ & ~ & $\checkmark$ & $\checkmark$  & 67.16 & 53.16 & 64.23 & 46.78 & 45.34\\
    \hline
\rowcolor[gray]{0.9} \textbf{Full model} & $\checkmark$ & $\checkmark$ & $\checkmark$ & $\checkmark$ & $\checkmark$   &\textbf{68.06} & \textbf{54.58} & \textbf{65.2} & \textbf{48.02} & \textbf{46.11}\\
    \bottomrule
    \end{tabular}
    }
     \caption{
    Ablation study on QVHighlights \textit{val} split. 
    A1-A4 analyze the modules of DRNet, while B1-B3 analyze the components in TCD. 
    CA, DK, and LGT denote cross-attention, dynamic kernels and learnable global tokens, respectively. 
    A5: replaces Mamba within CIOs with standard Transformer encoder.
    B2: replaces text-conditioned dynamic convolutions with standard convolutions. 
    B3: removes learnable global tokens used for global retrieval. }
    \label{ablation}
    \vspace{-3mm}
\end{table*}

\begin{figure*}[t]
    \centering
    \includegraphics[width=0.84\linewidth]{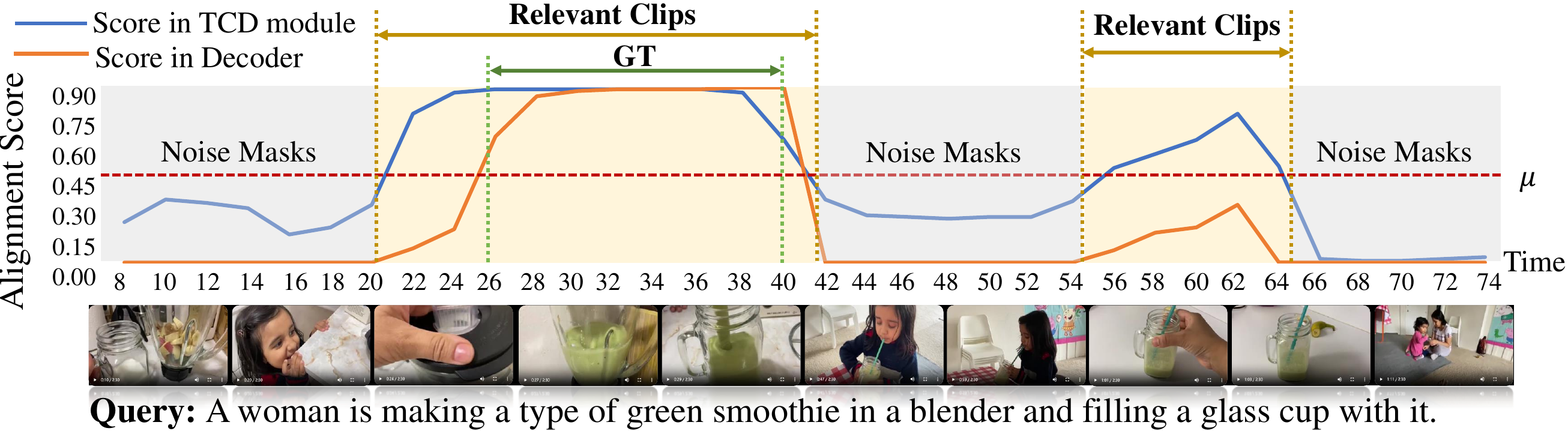}  
    % visualization-TRF.pdf
    \caption{Text-clip alignment scores show that our method effectively filters out noisy clips and accurately localizes the target moment within relevant segments.}
    % Comparison of text-clip alignment scores in text-conditioned denoising (TCD) and Masked Retrieval (MR) modules. Red dashed line is the noise filtering threshold \(\mu\) in TCD module, with gray background clips indicating those filtered out.}
    \label{fig_score}
    % \vspace{-2mm}
\end{figure*}

\begin{table*}[t]
\centering
\renewcommand{\arraystretch}{1.15}
    \addtolength{\tabcolsep}{0pt}
    \resizebox{0.65\textwidth}{!}{%
    \begin{tabular}{l|lllll}
    \hline
          Model & R1@0.5 &  R1@0.7 & mAP@0.5 & mAP@0.75 & mAP@Avg. \\
    \hline
         UniVTG\textsubscript{23iccv} &  60.52 & 42.39 & 59.08 & 37.12 & 36.66\\ 
         UniVTG\textsubscript{23iccv}$\dagger$ &  62.77 (\textcolor{red}{$\uparrow$ 2.25}) & 44.77 (\textcolor{red}{$\uparrow$ 2.38}) & 60.83 (\textcolor{red}{$\uparrow$ 1.75}) & 39.67 (\textcolor{red}{$\uparrow$ 2.55}) & 39.38 (\textcolor{red}{$\uparrow$ 2.72})\\ 
         VMS\textsubscript{24arxiv} & 65.48 & 50.06 & 62.92 & 45.2 & 43.62\\
         VMS\textsubscript{24arxiv}$\dagger$ &  66.58 (\textcolor{red}{$\uparrow$ 1.1}) & 51.87 (\textcolor{red}{$\uparrow$ 1.81}) & 64.51 (\textcolor{red}{$\uparrow$ 1.59}) & 45.97 (\textcolor{red}{$\uparrow$ 0.77}) & 44.76 (\textcolor{red}{$\uparrow$ 1.14})\\
    \hline               
         \rowcolor[gray]{0.9} \textbf{DRNet}  & \textbf{68.06} & \textbf{54.58} & \textbf{65.2} & \textbf{48.02} & \textbf{46.11}\\ 
         % \hline
          % UnLoc-B~\cite{yan2023unloc}~\pub{ICCV23} & 64.50 & 48.80 & - & - & -\\
    \bottomrule
    \end{tabular}
    }
    \caption{Comparison of methods with and without video denoising on QVHighlights \textit{val} split. All methods are re-implemented based on their official codes. $\dagger$ indicates results with denoising, with red values showing the performance improvement.}
    \label{generalization}
    % \vspace{-3mm}
\end{table*}

\paragraph{Datasets.} 
We validate the effectiveness of our method through extensive experiments on two popular datasets: QVHighlights and Charades-STA.
\textbf{QVHighlights}~\cite{lei2021detecting} is designed for moment retrieval and highlight detection, comprising over 10,000 videos, each averaging 150 seconds. The dataset includes 10,310 human-written text queries describing relevant segments, with an average segment length of 24.6 seconds, resulting in 18,367 annotated moments. We follow the original data splits, using the training set for model training and the test set for evaluation. 
\textbf{Charades-STA}~\cite{sigurdsson2016hollywood} is focused on temporal sentence grounding, derived from the Charades dataset. It contains 12,408 training and 3,720 testing moment-sentence pairs, with videos averaging 29.8 seconds in length, capturing various human actions and corresponding text queries.

% \vspace{-1mm}
\paragraph{Evaluation Metrics.}
Following previous VMR work~\cite{li2023g2l}, we use the standard evaluation metric R@n, IoU=m. This metric measures the percentage of queries that have at least one correctly retrieved moment (IoU \textgreater m) among the top-n output moments.
% We use the standard evaluation metrics `R@n, IoU=m'~\cite{li2023g2l, li2023d3g, yan2023unloc, lin2023univtg}, which calculate the percentage of queries with at least one correct retrieval (temporal IoU with the ground truth moment greater than m) in the top-n localized moments. 
For QVHighlights, we follow standard metrics~\cite{lei2020tvr}, using Recall@1 with IoU thresholds 0.5 and 0.7, mean average precision (mAP) with IoU thresholds 0.5 and 0.75, and the average mAP over a series of IoU thresholds [0.5:0.05:0.95] for moment retrieval. For Charades-STA, we follow~\cite{lin2023univtg} and use Recall@1 with IoU thresholds 0.3, 0.5, and 0.7, and mIoU.

% \vspace{-1mm}
\paragraph{Implementation Details.}
For a fair comparison, we use pre-extracted SlowFast and CLIP video features, and CLIP text features, for both datasets, provided by \cite{lin2023univtg}. In our DRNet, all encoders constructed using CIO consist of three CIO layers, each with a hidden size of $D = 1024$. Loss weights are set as: $\lambda_t=2$, $\lambda_{L1}^g=5$, $\lambda_{iou}^g=1$, $\lambda_{L1}^b=10$, $\lambda_{iou}^b=1$, and $\lambda_c=10$ for both datasets. For QVHighlights, $\lambda_{intra}$ and $\lambda_{inter}$ are set to 2 each, while for Charades-STA, they are set to 1 and 0.5, respectively. All experiments are conducted on a single RTX 3090 GPU.
% \vspace{-1mm}
% In all experiments, we use Adam~\cite{kingma2014adam} optimizer with 3e-5 learning rate and 1e-5 weight decay. The model is trained with batch size 32 for 200 epochs on QVHighlights, and batch size 8 for 100 epochs on Charades-STA. 

% \vspace{-5pt}
\subsection{Comparison to State-of-the-art Methods}
We compare our method to many state-of-the-art methods:
LLMEPET~\cite{jiang2024prior}, MomentDiff~\cite{li2024momentdiff}, MESM~\cite{liu2024towards}, UVCOM~\cite{xiao2024bridging}, LMR~\cite{liu2024context}, TR-DETR~\cite{sun2024tr},  VMS~\cite{chen2024video}, UniVTG~\cite{lin2023univtg}, QD-DETR~\cite{moon2023query}, UMT~\cite{liu2022umt}, RaNet~\cite{gao2021relation}, MomentDETR~\cite{lei2021detecting},MDETR~\cite{lei2021detecting}.
% 2D-TAN~\cite{zhang2020learning},  VSLNet~\cite{zhang2020span},
% MAN~\cite{zhang2019man}.
% MCN~\cite{anne2017localizing}, CAL~\cite{gao2017tall}.
% XML~\cite{lei2020tvr}, XML+~\cite{lei2020tvr}

% \vspace{2mm}
\noindent \textbf{QVHighlights.}
\cref{QVHighlights} compares our method with state-of-the-art (SOTA) approaches.
Our method sets new SOTA benchmarks, demonstrating significant improvements on all metrics. Specifically, it outperforms the latest 2024 methods by 9.63\% over MomentDiff\textsubscript{24neurips}, 3.64\% over MESM\textsubscript{24aaai}, 2.14\% over UVCOM\textsubscript{24cvpr}, and 1.4\% over TR-DETR\textsubscript{24aaai} on the average of all metrics. 
Notably, MomentDiff\textsubscript{24neurips} and UVCOM\textsubscript{24cvpr} use diffusion-based generative and general Transformer-based architectures, respectively, while MESM\textsubscript{24aaai} and TR-DETR\textsubscript{24aaai} employ DETR-based Transformer architectures. These substantial performance gains across various VMR architectures underscore the effectiveness and superiority of our method.
% CG-DETR~\cite{moon2023correlation} &  65.4 & 48.4 & \textbf{64.5} & 42.8 & 42.9\\

% \vspace{2mm}
\noindent \textbf{Charades-STA.} 
In \cref{Charades}, we evaluate our model's performance against the SOTA approaches using both VGG and SF+C backbones.
% we comprehensively evaluate the performance of our model against the latest SOTA models using both VGG and SF+C backbones. 
Our method consistently achieves top-tier performance across all metrics and backbones.
With the VGG backbone, our method outperforms MomentDiff\textsubscript{24neurips} and MESM\textsubscript{24aaai} by an average of 8.31\% and 2.16\%, respectively. For the SF+C backbone, our method surpasses the latest SOTA models by 4.76\% compared to TR-DETR\textsubscript{24aaai}, 2.56\% over LLMEPET\textsubscript{24arxiv}, and 2.38\% over UVCOM\textsubscript{24cvpr}.
In particular, for the challenging  mAP@0.7 and R1@0.7 metrics, which require high semantic alignment and IoU accuracy, our method surpasses the nearest competitors by 4.36\% and 3.15\% using the VGG and SF+C backbones, respectively.
Overall, the superior performance of our method across all metrics on real-world datasets demonstrates that removing noisy clips under text constraints and performing contextual fusion on the purified clips effectively enhances video moment retrieval (VMR).
% \subsection{4.3 \ Comparisons with Paradigm Transfer}
% \subsection{Experiments on General Architectures with QCDDL Paradigm}

\subsection{Ablation Study}
\paragraph{Ablation on DRNet.}\cref{ablation} (A1-A5) evaluate the contribution of each module in DRNet. A1-A3 present the results of removing the TCD, TRF, and Decoder modules respectively, each resulting in a performance drop. Notably, removing TCD (A1) leads to an average performance drop of 11.71\%, highlighting the critical role of noise filtering.
Compared with A3, removing both the TRF and Decoder modules in A4 results in a more significant performance drop than removing only the Decoder. This is because the semantic consistency loss computed in the TRF module provides auxiliary supervision for the denoising process, encouraging the generation of cleaner visual features for subsequent decoding.
% In Decoder, reconstructed text guides moment localization, and the losses between predicted and GT moment enhance text reconstruction. 
%Together with A1, we conclude that the three modules benefit each other and work synergistically to boost DRNet performance.

\noindent In A5, replacing Mamba with standard Transformer encoders as the base in DRNet leads to a significant 5.1\% performance drop. We attribute this to Mamba’s selective information propagation, which more effectively integrates key textual information into visual features than the Transformer’s global self-attention mechanism, which has information redundancy.

\noindent\paragraph{Ablation on TCD.}
Ablation experiments for removing each component in TCD (see B1-B3 of \cref{ablation}) shows a performance drop. The highest drop is observed for removing cross-attention (B1), highlighting the importance of cross-attention for integrating text and visual features after Mamba-based multimodal interaction. Similarly, the dynamic kernels (B2) and adding query tokens at the end of multimodal sequences also contribute to improved retrieval performance.

\noindent \paragraph{Ablation on CIO layers.}
In this ablation study, we investigate the effect of varying the layers of CIO in each module. Specifically, while adjusting the layers in one module, we keep the CIO layers in other modules fixed at three layers to ensure a fair comparison. As shown in \cref{abltaion_CIO}, using three layers of CIO consistently achieves optimal performance across all modules, confirming the effectiveness and simplicity of the current design.

\begin{table}[t]
\centering
\renewcommand{\arraystretch}{1.15}
    \addtolength{\tabcolsep}{0pt}
    % \vspace{-2mm}
    \resizebox{0.48\textwidth}{!}{
    \begin{tabular}{l|ccc|ccc|ccc}
    \toprule
          \multirow{2}{*}{Metrics} &  \multicolumn{3}{c|}{TCD} & \multicolumn{3}{c|}{TRF} & \multicolumn{3}{c}{Decoder} \\
    \cline{2-10}
          ~ & 2 & 3 & 4 & 2 & 3 & 4 & 2 & 3 & 4 \\
    \toprule
          R1@0.5 & 67.54 & 68.06 & 67.12 & 67.2 & 68.06 & 66.7 & 67.6 & 68.06 & 67.2 \\
          R1@0.7 & 54.25 & 54.58 & 53.96 & 53.8 & 54.58 & 53.5 & 54.1 & 54.58 & 53.5 \\
    \bottomrule
    \end{tabular}
    }
    \caption{Ablation study on CIO layers in  each module.} 
    \label{abltaion_CIO}
    \vspace{-3mm}
\end{table}

% \vspace{-1mm}
\noindent \paragraph{Denoise-then-Retrieve Paradigm Generalization.}
We apply the generated noise masks to existing Transformer-based (UniVTG) and Mamba-based (VMS) VMR methods without modifying their architectures. 
As shown in \cref{generalization}, this leads to notable performance gains across all metrics for both UniVTG~\cite{lin2023univtg} and VMS~\cite{chen2024video}. 
Specifically, by removing noisy clips, UniVTG and VMS see improvements of 2.72\% and 1.14\% points on mAP@Avg and 2.38\% and 1.81\% points on the challenging R1@0.7 metric, respectively.
These results underscore the effectiveness and generalization of our denoise-retrieve paradigm.

% UniVTG shows an improvement of 2.72\% in mAP@Avg, while VMS increases by 1.81\% on the challenging R1@0.7 metric. Our method still achieves the best performance on all metrics.
% show that our proposed paradigm is useful in general. 
%Overall, the effectiveness of denoising in both Transformer-based and Mamba-based methods demonstrates the generalization of the DRNet paradigm.

% mioabo multimodal represent \cite{miao2023spectrum}
% 24cvpr papers: BAM-DETR: Boundary-Aligned Moment Detection Transformer for Temporal Sentence Grounding in Videos

% \vspace{-2mm}
\subsection{Qualitative Results}
We present visualizations in \cref{fig_score}, where blue and orange curve denote text-clip alignment scores in TCD and Decoder modules, respectively. Red dashed line is the noise filtering threshold \(\mu\), with gray background clips indicating those filtered out.
The text-conditioned denoising module enables our method to effectively distinguish between text-relevant and noisy clips, while the decoder module can further localize the target clips within the text-relevant clips precisely.
Specifically, the blue curve illustrates that in TCD, there is a significant disparity in text-clip alignment scores between noisy and relevant clips, with consistently high alignment scores among the relevant clips. In contrast, the orange curve shows a clear distinction between the relevant clips, accurately identifying the text-referred clips. 
\section{Conclusion}
In this work, we analyzed the importance of denoising in Video Moment Retrieval (VMR) and introduced DRNet, a Text-conditioned Denoising and Text-reconstruction Feedback approach. Our method filters irrelevant clips by generating noise masks and refines the process by aligning re-generated queries, distilled from purified video representations, with the input text.  
Extensive experiments on Charades-STA and QVHighlights benchmarks validate the effectiveness of denoising, demonstrating substantial improvements over state-of-the-art methods and showcasing our paradigm's adaptability to enhance other VMR models.

% Our Text-Conditioned Denoising effectively filters out irrelevant video clips to generated noise masks.
% We then perform Text-Reconstruction Feedback to refine the denoising process by aligning the re-generated query and the language input.
% By effectively filtering out irrelevant video clips using our Text-Conditioned Denoising (TCD) module, and refining the process with Text-Reconstruction Feedback (TRF), we ensure purified multimodal representations that lead to significant performance improvement. 
% Extensive experiments on the Charades-STA and QVHighlights benchmarks confirm the effectiveness of denoising, demonstrating substantial improvements over existing state-of-the-art methods and showcasing the adaptability of our paradigm in enhancing other VMR models. 
% We hope that our denoise-then-retrieve paradigm will also benefit other video moment retrieval tasks in the future.

% \noindent \textbf{Limitation.} 
% We assess semantic consistency between re-generated query and input language embeddings using cosine similarity. Since videos can be expressed in different ways, more comprehensive metrics could better capture core semantic alignment to improve the denoising process.

% \newpage

\section*{Acknowledgments}
This work is supported by National Natural Science Foundation of China under Grants No.62472092, No. 62172089, No.62106045. Key Laboratory of Computer Network and Information Integration of Ministry of Education of China under Grants No. 93K-9, Nanjing Purple Mountain Laboratories, Fintech and Big Data Laboratory of Southeast University. We thank the Big Data Computing Center of Southeast University for providing the facility support on the numerical calculations.

\bibliographystyle{named}
\bibliography{ijcai25}

\end{document}